\documentclass{article}

\usepackage{arxiv}

\usepackage[utf8]{inputenc} 
\usepackage[T1]{fontenc}    
\usepackage{hyperref}       
\usepackage{url}            
\usepackage{booktabs}       
\usepackage{amsfonts}       
\usepackage{nicefrac}       
\usepackage{microtype}      
\usepackage{lipsum}
\usepackage{graphicx}

\usepackage{amsmath}
\usepackage{comment}
\usepackage{multirow}
\usepackage{makecell}
\usepackage{booktabs}
\usepackage{rotating}
\usepackage{dirtytalk}
\usepackage{xcolor}
\usepackage{hyperref}
\usepackage{array}
\usepackage{arydshln}
\usepackage{algpseudocode}
\usepackage{algorithm}

\title{Energy-Efficient Federated Learning via Adaptive Encoder Freezing for MRI-to-CT Conversion: A Green AI-Guided Research}

\author{
  Ciro Benito Raggio\thanks{Corresponding author} \\
  Institute of Biomedical Engineering\\
  Karlsruhe Institute of Technology\\
  Fritz-Haber-Weg 1, Karlsruhe 76131\\
  Baden-Württemberg, Germany \\
  \texttt{ciro.raggio@kit.edu} \\
  \And
  Lucia Migliorelli \\
  Department of Political Science\\
  Università Degli Studi Di Teramo\\
  Via Renato Balzarini 1, Teramo 64100\\
  Italy
  \And
  Nils Skupien \\
  Institute of Biomedical Engineering\\
  Karlsruhe Institute of Technology\\
  Fritz-Haber-Weg 1, Karlsruhe 76131\\
  Baden-Württemberg, Germany
  \And
  Mathias Krohmer Zabaleta \\
  Institute of Biomedical Engineering\\
  Karlsruhe Institute of Technology\\
  Fritz-Haber-Weg 1, Karlsruhe 76131\\
  Baden-Württemberg, Germany
  \And
  Oliver Blanck \\
  Department of Radiation Oncology \\
  University Medical Center Schleswig-Holstein\\
  Feldstrasse 21, Kiel 24105\\
  Schleswig-Holstein, Germany
  \And
  Francesco Cicone \\
  Department of Experimental and Clinical Medicine\\
  Magna Graecia University\\
  Viale Europa, Catanzaro 88100\\
  Italy
  \And
  Giuseppe Lucio Cascini \\
  Department of Experimental and Clinical Medicine\\
  Magna Graecia University\\
  Viale Europa, Catanzaro 88100\\
  Italy
  \And
  Paolo Zaffino \\
  Department of Experimental and Clinical Medicine\\
  Magna Graecia University\\
  Viale Europa, Catanzaro 88100\\
  Italy
  \And
  Maria Francesca Spadea \\
  Institute of Biomedical Engineering\\
  Karlsruhe Institute of Technology\\
  Fritz-Haber-Weg 1, Karlsruhe 76131\\
  Baden-Württemberg, Germany
}

\begin{document}
\maketitle
\begin{abstract}
\textbf{Background and Objective}: Federated Learning (FL) holds the potential to advance equality  in health by enabling diverse institutions to collaboratively train deep learning (DL) models, even with limited data. However, the significant resource requirements of FL often exclude centres with limited computational infrastructure, further widening existing healthcare disparities. To address this issue, we propose a Green AI-oriented adaptive layer-freezing strategy designed to reduce energy consumption and computational load while maintaining model performance.
\textbf{Methods}: We tested our approach using different federated architectures for Magnetic Resonance Imaging (MRI)-to-Computed Tomography (CT) conversion. The proposed adaptive strategy optimises the federated training by selectively freezing the encoder weights based on the monitored relative difference of the encoder weights from round to round. A patience-based mechanism ensures that freezing only occurs when updates remain consistently minimal. The energy consumption and $CO_{2eq}$ emissions of the federation were tracked using the CodeCarbon library. 
\textbf{Results}: Compared to equivalent non-frozen counterparts, our approach reduced training time, total energy consumption and $CO_{2eq}$ emissions  by up to 23\%. 
At the same time, the MRI-to-CT conversion performance was maintained, with only small variations in the Mean Absolute Error (MAE). Notably, for three out of the five evaluated architectures, no statistically significant differences were observed, while two architectures exhibited statistically significant improvements.
\textbf{Conclusions}: Our work aligns with a research paradigm that promotes DL-based frameworks meeting clinical requirements while ensuring climatic, social, and economic sustainability. It lays the groundwork for novel FL evaluation frameworks, advancing privacy, equity and, more broadly, justice in AI-driven healthcare.
\end{abstract}


\section{Introduction}
Healthcare is plagued by deep-rooted inequalities, with major disparities in access to services, infrastructure, and expertise across different regions and population groups \cite{macias2023race}. According to the World Health Organisation, about half of the world’s population lacks essential health services, underscoring the risk that new technologies could exacerbate these gaps if they remain financially out of reach \cite{world2023tracking}. 

Although Artificial Intelligence (AI) and particularly Deep Learning (DL) have shown great promise for supporting and improving diagnostic accuracy and patient's outcomes, their training and integration into routine care may require considerable computing power \cite{bolon2024review}. DL models with an increasing number of parameters and operations require large amounts of energy, in addition to water for cooling the data centres \cite{luccioni2024power, varoquaux2024hype}. Such resource-intensive approaches -- termed \textit{red AI} \cite{schwartz2020green} -- prioritise ever-higher performance at the cost of substantial computational and energy usage. This dynamic can reinforce barriers to access, restricting digital health to those who can afford it and undermining health equity. On the other side, \textit{green AI} (or, more recently, \textit{green-in AI} \cite{bolon2024review}) emphasises the integration of sustainable practices and techniques \textit{by design}, to train and deploy DL models in clinics while reducing the environmental costs and the financial burden \cite{schwartz2020green}. This paradigm advocates for assessing DL models not only on their quantitative performance, but also for their efficiency -- both in training and testing -- and therefore sustainability~\cite{liu2022trustworthy}. Consequently, in response to growing ethical and environmental concerns, the literature calls for more sustainable and socially responsible DL practices, requiring a reorientation of current research towards environmental and social sustainability as well as equity \cite{jobin2019global, raman2024green, siala2022shifting}. 

Complementing this, Federated Learning (FL) enables multiple institutions to collaboratively train DL models without centralizing data, preserving patient's privacy and broadening access to DL-based algorithms by potentially allowing even smaller centres to advance their diagnostic and research capabilities \cite{chen2025advances}.

Indeed, by participating in a federation, these centres can contribute with their data, thereby increasing data diversity and benefiting from a DL model that is potentially less prone to bias and designed to foster the development of more robust and generalisable models.

However, the effectiveness of FL is contingent upon the participating institutions' computational resources and infrastructure \cite{chen2025advances}. Centers with limited computational capabilities may struggle to engage in FL initiatives, potentially leading to a concentration of AI benefits among better-equipped institutions. This could result in a scenario where only well-resourced centers can fully leverage FL \cite{rauniyar2023federated}. 

Ensuring that FL does not exacerbate existing disparities or concentrate decision-making power within a few well-resourced institutions or countries, calls for a focus on equity, which entails actively reducing barriers that prevent resource-constrained institutions from fully participating \cite{narayanan2024fairness}. In response, inspired by recent contributions in the literature \cite{thakur2024green}, this work aims to lower FL’s computational demand by proposing an adaptive layers freezing method, to dynamically freeze the layers involved in training and aggregation, thereby mitigating computational overhead and making FL more accessible to resource-limited institutions. 

While prior research has explored layer freezing strategies to address heterogeneity in FL, existing works have primarily focused on communication efficiency and memory usage. For instance, \cite{chen2021communication, Malan2024} proposed a methodology for freezing parameters to reduce communication costs, and \cite{wu2024heterogeneity} introduced SmartFreeze to improve memory efficiency. However, these studies do not fully align their approaches with the principles of Green AI, particularly in the context of clinical applications and their social implications. By explicitly focusing on sustainability, this work contributes to making FL more energy-efficient and inclusive, facilitating wider participation without compromising the federated model's performance.

For our purposes, we elected as our case study the federated inter-modality medical Image-to-Image (I2I) translation task. Inter-modality I2I translation converts data between different image modalities (e.g., Magnetic Resonance Imaging (MRI) to Computed Tomography (CT)), offering benefits in areas like radiotherapy planning, by enhancing treatment accuracy and reducing additional radiation \cite{thummerer2024artificial}.

Our recent study introduced FedSynthCT-Brain \cite{raggio2025fedsynthct}, a FL framework built upon encoder-decoder architectures, such as U-Net variants. The work dealt with a benchmark analysis aimed at identifying architectures that are both effective and computationally efficient for the MRI-to-CT synthesis task. It emphasized the need to further optimize computational efficiency as a natural direction for future research. Thus, this study builds on our FedSynthCT-Brain \cite{raggio2025fedsynthct} and is guided by research paradigms that advocate for DL-based clinical decision support systems (CDSS) that, \textit{by design}, meet clinical requirements while upholding a commitment to optimization, sustainability and equity \cite{raman2024green}. 

We specifically introduce a patience-based mechanism, also inspired by early stopping strategies commonly used in DL \cite{Prechelt2012}, combined with the overall distance between consecutive encoder weights update. The patience mechanism is crucial in ensuring that freezing occurs exclusively when the designated network components (i.e., the encoder) achieved the stability. This facilitates the comprehension of the client consensus, as the aggregated model weights' distance undergoes a natural reduction over successive rounds, culminating in a synchronised approach across all participants.

\section{Related work} \label{sec:related_works}

The application of FL to medical I2I translation is a recent development, with only a few studies specifically addressing this task \cite{WANG2023126282, raggio2025privacypreservingfederatedlearningframework}. Among them is our recent FedSynthCT-Brain framework \cite{raggio2025fedsynthct}, which tackles MRI-to-CT translation in a federated setting using a cross-silo horizontal FL approach. In this previous work, we observed that the task’s complexity and the high resolution of medical images pose a significant computational burden on individual clients, leading to extended training times.

A common initial optimisation strategy to address the aforementioned issue is to reduce communication overhead, which primarily arises from frequent synchronisation between the central server and clients during weight exchanges. A typical solution is to increase the number of local training epochs, thereby reducing synchronisation frequency. While this reduces communication costs, it may come at the expense of model accuracy, as more frequent synchronisations have been shown to improve knowledge sharing among clients and thus model performance \cite{pmlr-v54-mcmahan17a}. The trade-off analysis in \cite{pmlr-v54-mcmahan17a} suggests that a moderate increase in local epochs can reduce communication costs with only marginal accuracy degradation.

The implementation of alternative optimisation strategies is focused on the objective of reducing bandwidth usage, through the application of model compression techniques, such as pruning, quantisation and their combinations, such as FedWSOcomp \cite{Mills2020,Nguyen2023, Onaizah2025}. However, these approaches result in a substantial increase in resource consumption, due to the additional computational load associated with the model compression and decompression on both client and server sides \cite{Malan2024}.

Several studies have also focused on developing customised aggregation functions to mitigate the adverse effects of non-IID and imbalanced data, thereby improving convergence and reducing the number of required communication rounds. One potential solution is the incorporation of a regularisation term within FedProx \cite{LiFedProx2020}, which serves to penalise local updates that deviate from the global model. In our previous study, we demonstrate that the employment of FedAvg in combination with FedProx achieves optimal performance, with FedProx typically requiring fewer federated rounds when compared with other strategies such as FedAvg, FedAvgM, FedBN and FedYogi \cite{raggio2025fedsynthct}. However, this optimisation proved insufficient to reduce the local computational burden or training time to a noteworthy extent, indicating that further optimisation of local training is still necessary.

In the context of federated I2I translation tasks, where encoder-decoder networks (such as U-Net, residual U-Net, or as generators in GAN architectures) are frequently employed  \cite{SpadeaMaspero2021, DAYARATHNA2024103046, HUIJBEN2024103276}, the methodology of freezing layers emerged as a particularly relevant area of interest. This approach demonstrates to be an effective training optimisation technique, both in centralised and federated settings \cite{Goutam2020, pfeiffer2023cocoflcommunicationcomputationawarefederated, MalanGradualFreezing2023, Malan2024}. In the context of centralised training, layer freezing demonstrates to  accelerate the training process by circumventing the back-propagation procedure for the frozen layers. In federated settings, the use of freezing may reduce communication costs and the computational load on individual clients.

The most common freezing strategies involve \say{layer-wise} freezing, where specific layers are held constant throughout training. Over time, this reduces the computational load by skipping updates for the frozen parameters. 

Both adaptive and permanent layer freezing approaches have been explored. In adaptive freezing, layers may be frozen or unfrozen at different training iterations, based either on random selection or guided by proxies, such as the magnitude of weight updates across epochs or classification accuracy on a calibration set. In permanent freezing, once a layer is frozen, it remains unchanged until training concludes \cite{Malan2024}. 

Several studies have applied adaptive layer freezing to reduce local computational costs and encourage client participation \cite{Goutam2020, pfeiffer2023cocoflcommunicationcomputationawarefederated, MalanGradualFreezing2023, Malan2024}. FedPT \cite{sidahmed2021efficientprivatefederatedlearningFEDPT} proposes reducing communication costs by statically selecting a subset of layers to train throughout the entire learning process, while keeping the remaining layers fixed with randomly initialised weights. However, this manual selection of trainable layers cannot guarantee optimality and may lead to substantial accuracy losses \cite{Malan2024}. Furthermore, other works propose incremental freezing schedules based on the topological order of network layers, following predefined manual schedules or introduce an automated procedure for selecting layers to freeze by monitoring the evolution of client-model updates \cite{MalanGradualFreezing2023, Malan2024}.

The method proposed in this paper leverages the benefits of freezing in federated training, while adapting it to different encoder-decoder architectures used in I2I translation tasks~\cite{SpadeaMaspero2021, DAYARATHNA2024103046, HUIJBEN2024103276}.

Despite recent advancements, none of the state-of-the-art works have explored freezing strategies in the context of federated medical I2I translation (such as the MRI-to-CT task tackled in FedSynthCT-Brain \cite{raggio2025fedsynthct}). Furthermore, our method addresses key limitations highlighted in previous works, including the study by Malan et al. \cite{Malan2024}. Although their approach has shown promising results in cross-device federated classification using shallow CNNs, applying it to deeper models poses scalability challenges. 

In contrast, our method introduces a patience-based adaptive freezing strategy that dynamically determines when to freeze encoder layers based on a quantitative measure of inter-round convergence in aggregated encoder weights. The mechanism incorporates user-controlled stability criteria --such as the patience rounds-- to distinguish between transient fluctuations and stable convergence, ensuring robustness and scalability to complex encoder–decoder architectures. Our methodology is designed in line with the principles of Green AI \cite{schwartz2020green}, prioritizing computational efficiency as a way to reach federated sustainability. By reducing client-side training loads via selective freezing, we aim to limit energy consumption and promote more responsible AI practices.

\section{Methods}

\subsection{Federeated Learning Setup and Architectures}
\label{subsec:fl_setup}

We designed a FL environment to simulate a realistic collaboration among four distinct clinical centres, in accordance with the setup employed in the original study \cite{raggio2025fedsynthct}. Each centre trained a local DL model on its own data, while a central server coordinated the federated model by aggregating key parameters (e.g., weights, biases, batch normalization) from all clients after each training round. Following the findings outlined in \cite{raggio2025fedsynthct}, we used as DL models for our FL framework:

\begin{itemize}
    \item The original U-Net \cite{UNetRonnenberger}, referred to as Simple U-Net, featuring an encoder-decoder structure with concatenation-based long skip connections;
    
    \item The architecture proposed by Li et al. \cite{Li2019}, designed specifically for the MRI-to-CT translation task, characterised by an encoder-decoder structure with residual connections instead of traditional skip connections, and identified as the best trade-off between effectiveness and efficiency according to~\cite{raggio2025fedsynthct};
    
    \item A deep encoder-decoder network inspired by U-Net, proposed by Spadea, Pileggi et al. \cite{SPADEA2019495}, which is distinguished by its increased depth, extensive use of multi-layer convolutional blocks at each level, and regularisation via dropout;
    
    \item The architecture introduced by Fu et al. \cite{FuetalUNet2019} for synthetic CT generation from MRI in the male pelvic region, which adopts a U-Net-like structure, but differs in its deep bottleneck, the use of instance normalisation, residual additions instead of concatenations, and the exclusive reliance on transposed convolutions for upsampling;
    
    \item A lightweight U-Net variant proposed by Li et al. \cite{LiUNet2020} for the MRI-to-CT translation task, characterised by the use of Leaky ReLU activations and concatenation-based skip connections.
\end{itemize}

All the architectures predicted $256\times256$ synthetic CT (sCT) slices from $256\times256$ MRI inputs.

\label{sec:Datasets}
\subsection{Datasets}

As for the original study \cite{raggio2025fedsynthct}, the experiments were conducted using the aforementioned FL setup with 4 clients (Centre A, B, C, and D) and 1 server (Centre E) on brain images.

\begin{table}[tbp!]
\centering
\small
\caption{Dataset key characteristics across participating centres.}
\label{tab:datasets}
\setlength{\tabcolsep}{0.008\linewidth}
\begin{tabular}{cccccc}
\toprule
\textbf{Centre} & \textbf{Data} & \textbf{\makecell{MR Voxel\\
{[}mm$^{3}${]}}} & \textbf{\makecell{CT Voxel\\
{[}mm$^{3}${]}}} & \makecell{\textbf{Size}\\\textbf{(Axial$\times$Sagittal$\times$Coronal)}}\\
\midrule
\textbf{A} & 15 & 1$\times$1$\times$1 & \makecell[c]{0.49-0.67 $\times$\\0.49-0.67 $\times$\\ 2.5} & 256$\times$176$\times$256\\
\midrule
\textbf{B} & 14 & 1$\times$1$\times$1 & 0.98$\times$0.98$\times$3.27 & 256$\times$176$\times$248\\
\midrule
\textbf{C} & 21 & 0.78$\times$0.78$\times$1 & 0.78$\times$0.78$\times$1 & \makecell[c]{226-357 $\times$\\512$\times$512}\\
\midrule
\textbf{D} & 29 & \makecell[c]{0.98-1.12 $\times$\\0.98-1.12 $\times$\\0.98-1.12} & \makecell[c]{0.69-0.78 $\times$\\ 0.69-0.79 $\times$\\ 1-3} & \makecell[c]{167-213 $\times$\\216-262 $\times$\\250-277}\\
\midrule
\textbf{E} & 23 & 0.98$\times$0.98$\times$3.27 & \makecell[c]{0.59-1.27 $\times$\\ 0.59-1.27 $\times$\\ 1-2} & \makecell[c]{167-262 $\times$\\200-225 $\times$\\225-248}\\
\bottomrule
\end{tabular}
\end{table}

Centres A, B, and C contained private datasets from US, Italy and Germany respectively, acquired in compliance with the ethical standards of the 1964 Declaration of Helsinki and its later amendments, with written informed consent obtained from all patients. Centres D and E were single institutions selected from the SynthRAD Grand Challenge 2023 \cite{synthrad23}.  Image dimensions and number of the samples in each dataset are detailed in Table \ref{tab:datasets}. Further details on the scanners used for image acquisition for each centre were reported in the previous study \cite{raggio2025fedsynthct}.

Each client represented a distinct hospital (or silo) with unique and non-overlapping datasets. The federated model was further validated on the unseen dataset of the server (Centre E), which was not included in any client’s training and validation process. This setup ensured that the server-side evaluation process tested the generalisation capabilities of the federated model on an unseen dataset with different characteristics.

Following recent findings \cite{raggio2025fedsynthct}, \cite{Han2025} a pre-preprocessing pipeline was applied to harmonize the datasets without exchanging data. Firstly, N4 bias field correction was applied with different parameters for each client. All MRIs were then rescaled and normalised to the [0,1] range. To ensure uniform input dimensions for the networks, all images were resized to 256$\times$256$\times$256 using a combination of cropping, resizing, and padding techniques. Additionally, each client applied data augmentation using spatial transformations, such as rotations, translations, and flipping.

\subsection{Patience-based Adaptive Layer Freezing}

\begin{figure}[tbp!]
\centering
\includegraphics[width=\linewidth]{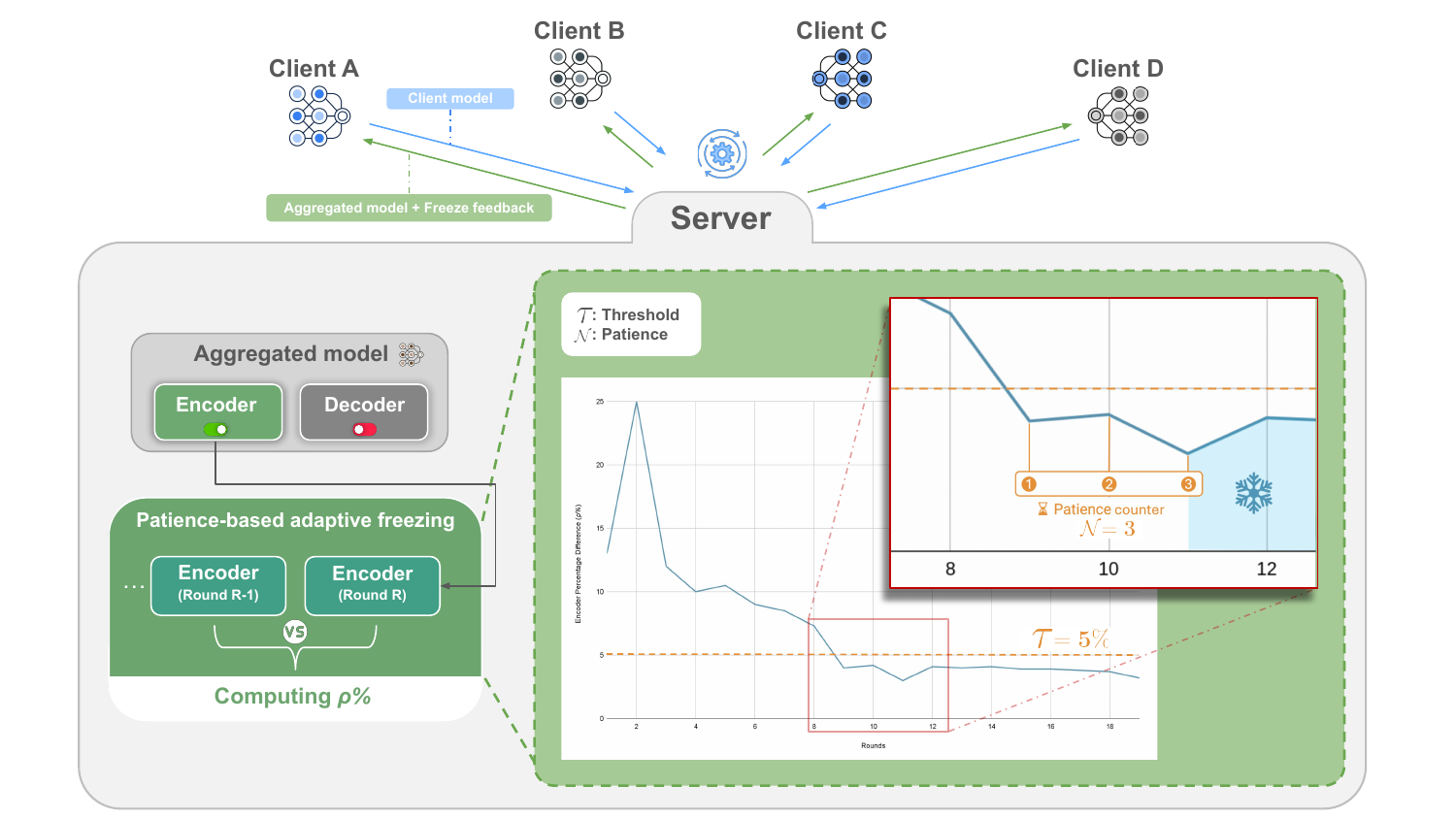}
\caption{Representation of the proposed patient-based adaptive layer freezing. The server is responsible for calculating the percentage difference in the aggregated encoder ($\rho_{\%}$) round by round. The encoder freezing condition is triggered when $\rho _{\%}$ remains consistently below the threshold ($\tau$) established by the final user for a minimum of $\mathcal{N}$ (patience parameter) consecutive rounds.} 
\label{fig:method}
\end{figure}

The concept of the proposed method is to selectively freeze the encoder weights of encoder-decoder architectures. This strategy aims to lessen the computational requirements within the FL framework, while preserving the performance akin to those of the unfrozen architectures (i.e., encoder-decoder based models without layer freezing).

As presented in Figure \ref{fig:method}, the proposed method quantified the alignment of clients' models during federated training by tracking incremental changes in encoder weights across consecutive training rounds. The key metric of the proposed method was the \textit{relative percentage difference} (\( \rho_{\%} \)) of encoder weights, computed as:
\begin{equation}
     \rho_{\% round} = \frac{f(W_{\text{R}}) - f(W_{\text{R-1}})}{f(W_{\text{R}})} \times 100
\end{equation}

Where: \( W_{\text{R}} \) represented the mean encoder weights at the current training round, \( W_{\text{R-1}} \) represented the mean encoder weights from the previous training round and \( f(\cdot) \) computed the Mean Absolute Error (MAE) between corresponding encoder layers.  

The MAE is defined as:
\begin{equation}
    \text{MAE}_{round} = \frac{1}{n} \sum_{i=1}^{n} \left| L_{\text{R}}^{(i)} - L_{\text{R-1}}^{(i)} \right|
\end{equation}

Where \( L_{\text{R}}^{(i)} \) and \( L_{\text{R-1}}^{(i)} \) denote the weights of the \( i \)-th encoder layer for the current and previous training rounds, respectively, and \( n \) was the total number of layers. The average MAE across layers provided a measure of weight divergence.

As clients optimised the global objective rather than their local objectives round after round, the \( \rho_{\%} \) decreased over successive training rounds (see Figure \ref{fig:method}), serving as an indicator of clients' consensus in the federated task. \newline

A \emph{patience mechanism} was then introduced to manage encoder freezing based on \( \rho_{\%} \). A minimum threshold $\tau$ was established to distinguish values of \( \rho_{\%} \) from minor fluctuations. The patience parameter \( \mathcal{N} \) was defined to specify the number of consecutive rounds during which \( \rho_{\%} \) must remain below $\tau$ to trigger the mechanism. The patience mechanism is formalised as:

\begin{equation}
    \text{Freeze Federated Encoder if: } \sum_{t=R-\mathcal{N}+1}^{R} \mathcal{I}(\rho_{\%}^{(t)} < \tau) = \mathcal{N}
\end{equation}

Where, $R$ is the current round, $\mathcal{N}$ is the patience parameter, defining the number of consecutive rounds where $\rho_{\%}<\tau$, $\mathcal{I}(\cdot)$ is the indicator function, returning 1 if the condition is true, and 0 otherwise, $\rho_{\%}^{(t)}$ is the percentage difference at round $t$, $\tau$ is the minimum threshold for $\rho_{\%}$ to be considered converged. The $t$ was defined as $R-N+1$. Thus, the condition was only satisfied in the last $N$ consecutive rounds. Once $\rho_{\%}$ was below the threshold $\tau$ for the defined $\mathcal{N}$ consecutive rounds, the encoder was frozen, which indicates sufficient convergence for the encoder. 

Figure \ref{fig:method} provides a conceptual illustration of the aforementioned patience-based freezing mechanism, with a specified threshold of $\tau=5\%$ and a patience of $\mathcal{N}=3$ rounds. The server monitors \( \rho_{\%} \) at each round, and once the freezing condition is satisfied, it notifies the clients to freeze the encoder.

\subsection{Experimental settings and performance metrics}
For our purposes, we used the experimental setup proposed in~\cite{raggio2025fedsynthct}, which used PyTorch~\cite{paszke2019pytorch} and MONAI~\cite{cardoso2022monaiopensource} for the DL implementation, and the Flower framework \cite{beutel2022flowerfriendlyfederatedlearning} for the FL implementation. 

All experiments were conducted in a containerised environment equipped with 32GB of RAM, 1 NVIDIA A100 80GB GPU, and 16 CPU cores. These computational resources were required to support the load of the entire federation (see Section \ref{subsec:fl_setup}) within a single physical node. Each client was nevertheless treated as an independent remote centre, and no individual client would require comparable resources in an actual distributed deployment.

The data augmentation process was implemented using the AugmentedDataLoader library \footnote{https://github.com/ciroraggio/AugmentedDataLoader}. The CodeCarbon\footnote{https://codecarbon.io/} \cite{Bouza_2023} library was used to track the emissions and the consumption of each federated training \cite{schwartz2020green}.

To ensure reproducibility, each experiment was repeated five times using an identical setup for each client: a batch size of 8, a learning rate of $10^{-4}$, 1 epoch of training for each local client for each round and the Adam optimizer. Following the experimental protocol and findings of our previous work \cite{raggio2025fedsynthct}, models were trained using the Random Multi-2D approach, and a voting strategy to reach a consensus on the predictions made for each anatomical plane was applied before the test; the best performing aggregation strategy, FedAvg combined with FedProx (with a proximal term of 3), was used for each experiment. Finally, as convergence and stability were consistently achieved within 25 communication rounds and no substantial performance improvements were observed beyond this~\cite{raggio2025fedsynthct}, a fixed number of 25 rounds was used in all experiments. Therefore, a fixed number of 25 rounds was used in all experiments.

To validate our approach, we computed a set of metrics, including time, emissions, and energy consumption, both with and without the proposed methodology, in an isolated environment.

To assess the impact of the methodology on image similarity, we employed four datasets (Centres A, B, C, and D) during the federation phase and an additional external dataset (Centre E) to evaluate the federated model’s generalisation performance on previously unseen data. Image similarity was assessed using well-established metrics for the MRI-to-CT translation task \cite{SpadeaMaspero2021, DAYARATHNA2024103046, raggio2025fedsynthct}, such as: (i) the Mean Absolute Error (MAE); (ii) the Peak Signal-to-Noise Ratio (PSNR); and (iii) the Structural Similarity Index Measure (SSIM), computed between the ground truth CT and the predicted CT.

\subsection{Observational Study and Adaptive Freezing Evaluation}

In the first set of experiments, a preliminary study was performed. Models were trained without the encoder freezing in the FL setup. The goal was to observe the update trend of the encoder weights across training rounds and establish baseline patterns for subsequent studies.

Two percentage difference thresholds were selected for the observational study, specifically $\tau = 5\%$ and $\tau = 10\%$. Both values were tested using a fixed number of patience rounds ($\mathcal{N}=3$). 

Analysis of the encoder weight relative difference trends indicated that the 10\% threshold was consistently reached after approximately 5–6 rounds, whereas the 5\% threshold was typically reached after 7–10 rounds. 

It was observed that, although the activation of the freezing mechanism was delayed when using the more conservative $\tau = 5\%$, model performance remained unchanged while the time needed to complete the rounds was considerably reduced. 

Nevertheless, employing the higher threshold ($\tau = 10\%$) could have yielded further gains in terms of training time, energy consumption, and emissions. However, this configuration resulted in performance degradation across different models, due to the anticipation of encoder freezing, thereby undermining the reliability of the approach.

Following these findings, using the same setup, the models were trained for 25 rounds across 5 repetitions, with the patience parameter (\( \mathcal{N} \)) set to 3 rounds to ensure that the threshold ($\tau=5\%$) was consistently exceeded.

The final set of experiments validated the proposed methodology by comparing model performance in scenarios with and without adaptive encoder freezing.

\section{Results}
A detailed comparison of the average emissions, energy consumption, time performance and number of trainable parameters across five independent repetitions for each evaluated model, both with and without the proposed methodology, is presented in Table \ref{tab:comparative_results}. 

\begin{table}[ht]
    \centering
    \caption{Results (mean $\pm$ standard deviation) obtained from 5 repetitions for each model and methodology.}
    \label{tab:comparative_results}
    \begin{tabular}{llcc}
        \toprule
        \textbf{Model} & \textbf{Metric} & \textbf{\makecell{No Adaptive\\Freeze}} & \textbf{\makecell{Adaptive Freeze\\$\mathcal{N}=3, \space \tau=5\%$}} \\
        \midrule
        \multirow{4}{*}{Simple U-Net \cite{UNetRonnenberger}} 
        & Parameters [\#] & $1.727 \times 10^{7}$ & $7.857 \times 10^{6}$ \\
        & Time [h] & 13.40 $\pm$ 0.76 & \textbf{11.15} $\pm$ \textbf{0.64} \\
        & Emissions [kgCO\textsubscript{2}eq] & 2.03 $\pm$ 0.09 & \textbf{1.69} $\pm$ \textbf{0.09} \\
        & Total energy [kWh] & 5.32 $\pm$ 0.23 & \textbf{4.45} $\pm$ \textbf{0.23} \\
        \midrule
        \multirow{4}{*}{Li et al. \cite{Li2019}} 
        & Parameters [\#] & $2.544 \times 10^{7}$ & $1.272 \times 10^{7}$ \\
        & Time [h] & 10.11 $\pm$ 0.38 & \textbf{7.89} $\pm$ \textbf{0.22} \\
        & Emissions [kgCO\textsubscript{2}eq] & 1.51 $\pm$ 0.04 & \textbf{1.16} $\pm$ \textbf{0.03} \\
        & Total energy [kWh] & 3.96 $\pm$ 0.10 & \textbf{3.05} $\pm$ \textbf{0.07} \\
        \midrule
        \multirow{4}{*}{Spadea, Pileggi et al. \cite{SPADEA2019495}} 
        & Parameters [\#] & $4.791 \times 10^{7}$ & $3.065 \times 10^{7}$ \\
        & Time [h] & 19.22 $\pm$ 0.75 & \textbf{17.42} $\pm$ \textbf{0.68} \\
        & Emissions [kgCO\textsubscript{2}eq] & 2.97 $\pm$ 0.10 & \textbf{2.70} $\pm$ \textbf{0.10} \\
        & Total energy [kWh] & 7.80 $\pm$ 0.26 & \textbf{7.08} $\pm$ \textbf{0.27} \\
        \midrule
        \multirow{4}{*}{Fu et al. \cite{FuetalUNet2019}} 
        & Parameters [\#] & $3.492 \times 10^{7}$ & $2.729 \times 10^{7}$ \\
        & Time [h] & 16.49 $\pm$ 0.57 & \textbf{13.81} $\pm$ \textbf{0.64} \\
        & Emissions [kgCO\textsubscript{2}eq] & 2.17 $\pm$ 0.07 & \textbf{1.80} $\pm$ \textbf{0.07} \\
        & Total energy [kWh] & 5.69 $\pm$ 0.17 & \textbf{4.74} $\pm$ \textbf{0.20} \\
        \midrule
        \multirow{4}{*}{Li et al, 2020 \cite{LiUNet2020}} 
        & Parameters [\#] & $2.682 \times 10^{7}$ & $9.024 \times 10^{6}$ \\
        & Time [h] & 14.10 $\pm$ 0.28 & \textbf{11.76} $\pm$ \textbf{0.59} \\
        & Emissions [kgCO\textsubscript{2}eq] & 1.85 $\pm$ 0.03 & \textbf{1.54} $\pm$ \textbf{0.07} \\
        & Total energy [kWh] & 4.85 $\pm$ 0.09 & \textbf{4.04} $\pm$ \textbf{0.19} \\
        \bottomrule
    \end{tabular}
\end{table}

\begin{table}[tbp]
\centering
\caption{Training time (in minutes) per local epoch (mean $\pm$ standard deviation) before and after adaptive encoder freezing for each centre and model.}
\begin{tabular}{llcc}
\toprule
\textbf{Centre} & \textbf{Model} & \multicolumn{2}{c}{\textbf{Training time per local epoch [min]}} \\
\cmidrule(lr){3-4}
 & & Pre-freeze & Post-freeze \\
\midrule
\multirow{5}{*}{Centre A} 
 & Simple U-Net \cite{UNetRonnenberger}           & 21.07 $\pm$ 0.65 & \textbf{14.38 $\pm$ 0.43} \\
 & Li et al. \cite{Li2019}                        & 15.59 $\pm$ 0.53 & \textbf{10.05 $\pm$ 0.31} \\
 & Spadea, Pileggi et al. \cite{SPADEA2019495}   & 31.35 $\pm$ 1.04 & \textbf{25.36 $\pm$ 0.80} \\
 & Fu et al. \cite{FuetalUNet2019}               & 26.29 $\pm$ 2.02 & \textbf{18.81 $\pm$ 1.11} \\
 & Li et al. 2020 \cite{LiUNet2020}              & 21.37 $\pm$ 0.47 & \textbf{15.14 $\pm$ 0.33} \\
\midrule
\multirow{5}{*}{Centre B} 
 & Simple U-Net \cite{UNetRonnenberger}           & 20.12 $\pm$ 1.60 & \textbf{13.52 $\pm$ 0.90} \\
 & Li et al. \cite{Li2019}                        & 14.79 $\pm$ 1.07 & \textbf{9.46 $\pm$ 0.64} \\
 & Spadea, Pileggi et al. \cite{SPADEA2019495}   & 29.34 $\pm$ 1.68 & \textbf{23.71 $\pm$ 1.26} \\
 & Fu et al. \cite{FuetalUNet2019}               & 25.60 $\pm$ 2.13 & \textbf{18.25 $\pm$ 1.56} \\
 & Li et al. 2020 \cite{LiUNet2020}              & 19.85 $\pm$ 0.49 & \textbf{14.02 $\pm$ 0.34} \\
\midrule
\multirow{5}{*}{Centre C} 
 & Simple U-Net \cite{UNetRonnenberger}           & 28.38 $\pm$ 1.87 & \textbf{19.31 $\pm$ 1.03} \\
 & Li et al. \cite{Li2019}                        & 20.15 $\pm$ 0.44 & \textbf{13.32 $\pm$ 0.25} \\
 & Spadea, Pileggi et al. \cite{SPADEA2019495}   & 39.77 $\pm$ 0.66 & \textbf{32.41 $\pm$ 0.49} \\
 & Fu et al. \cite{FuetalUNet2019}               & 35.80 $\pm$ 1.75 & \textbf{25.78 $\pm$ 1.21} \\
 & Li et al. 2020 \cite{LiUNet2020}              & 29.29 $\pm$ 1.67 & \textbf{20.89 $\pm$ 1.20} \\
\midrule
\multirow{5}{*}{Centre D} 
 & Simple U-Net \cite{UNetRonnenberger}           & 30.54 $\pm$ 1.40 & \textbf{20.76 $\pm$ 0.82} \\
 & Li et al. \cite{Li2019}                        & 22.03 $\pm$ 0.58 & \textbf{14.49 $\pm$ 0.36} \\
 & Spadea, Pileggi et al. \cite{SPADEA2019495}   & 44.03 $\pm$ 0.79 & \textbf{35.77 $\pm$ 0.59} \\
 & Fu et al. \cite{FuetalUNet2019}               & 38.50 $\pm$ 1.90 & \textbf{27.71 $\pm$ 1.35} \\
 & Li et al. 2020 \cite{LiUNet2020}              & 31.72 $\pm$ 1.41 & \textbf{22.57 $\pm$ 1.06} \\
\bottomrule
\end{tabular}
\label{tab:training_times}
\end{table}

\begin{figure}[tbp!]
\centering
\includegraphics[width=\linewidth]{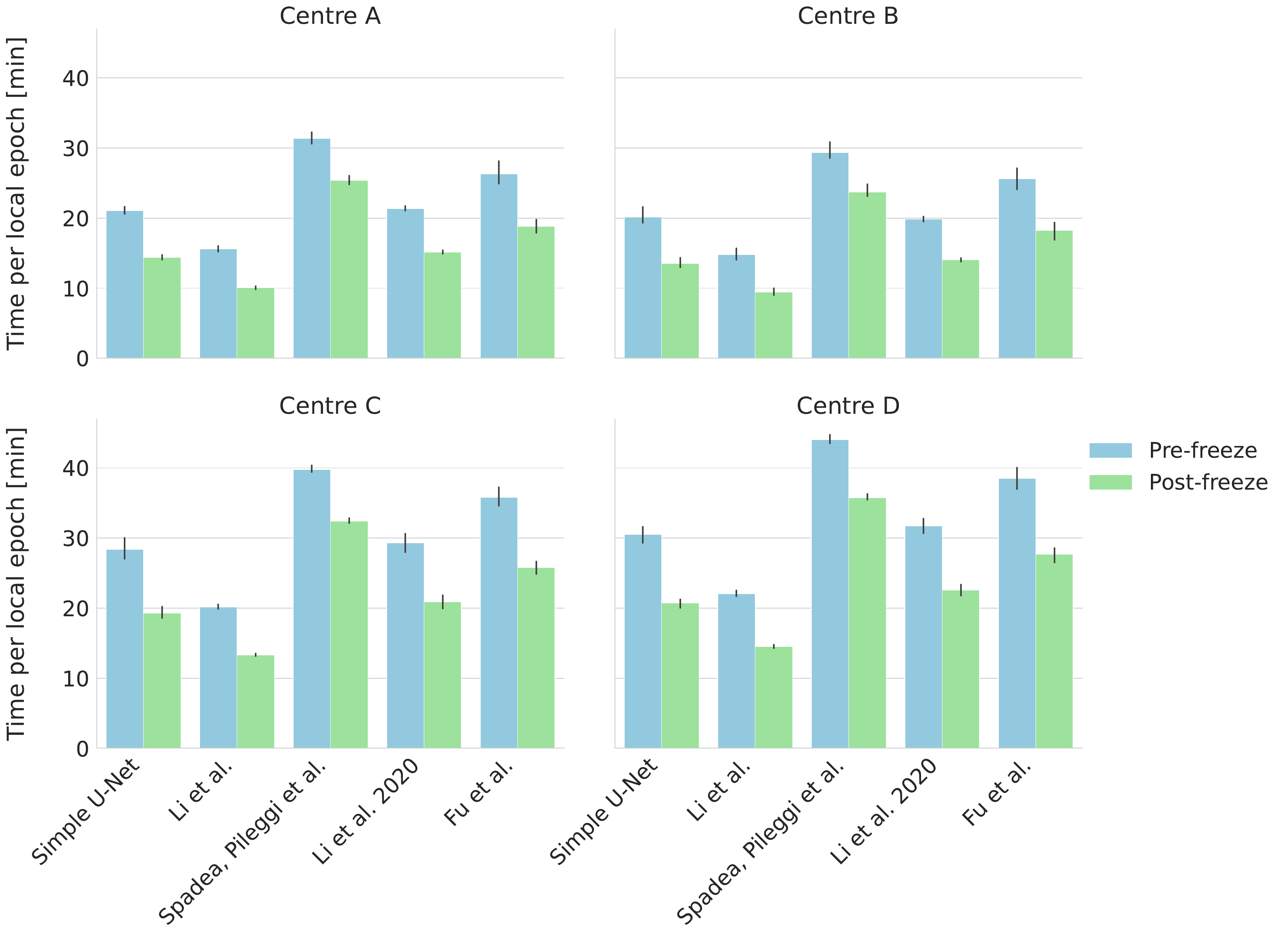}
\caption{Training time per local epoch for each participating client (Centres A-D), comparing the pre-freeze and post-freeze phases across the evaluated models.} \label{fig:time_client_comp_pre_post}
\end{figure}

A consistent reduction in the number of trainable parameters was observed following the activation of the encoder freezing controlled using the proposed methodology. Therefore, a reduction in training time per local epoch was observed on all clients (see Figure \ref{fig:time_client_comp_pre_post} and Table \ref{tab:training_times}), leading to a global decrease in the total training time across all architectures, with improvements ranging from $9.4\%$ (Spadea, Pileggi et al. \cite{SPADEA2019495}) to $22.0\%$ (Li et al. \cite{Li2019}). 

These temporal gains translated into proportional decreases in total energy consumption and CO\textsubscript{2}eq emissions, with reductions between $9.1\%$ and $23.2\%$. The Li et al. \cite{Li2019} model showed the largest benefits, while the Simple U-Net \cite{UNetRonnenberger}, Fu et al. \cite{FuetalUNet2019}, and Li et al. 2020 \cite{LiUNet2020} achieved consistent improvements around $16$–$17\%$. 

\begin{figure}[tbp!]
\centering
\includegraphics[scale=.5]{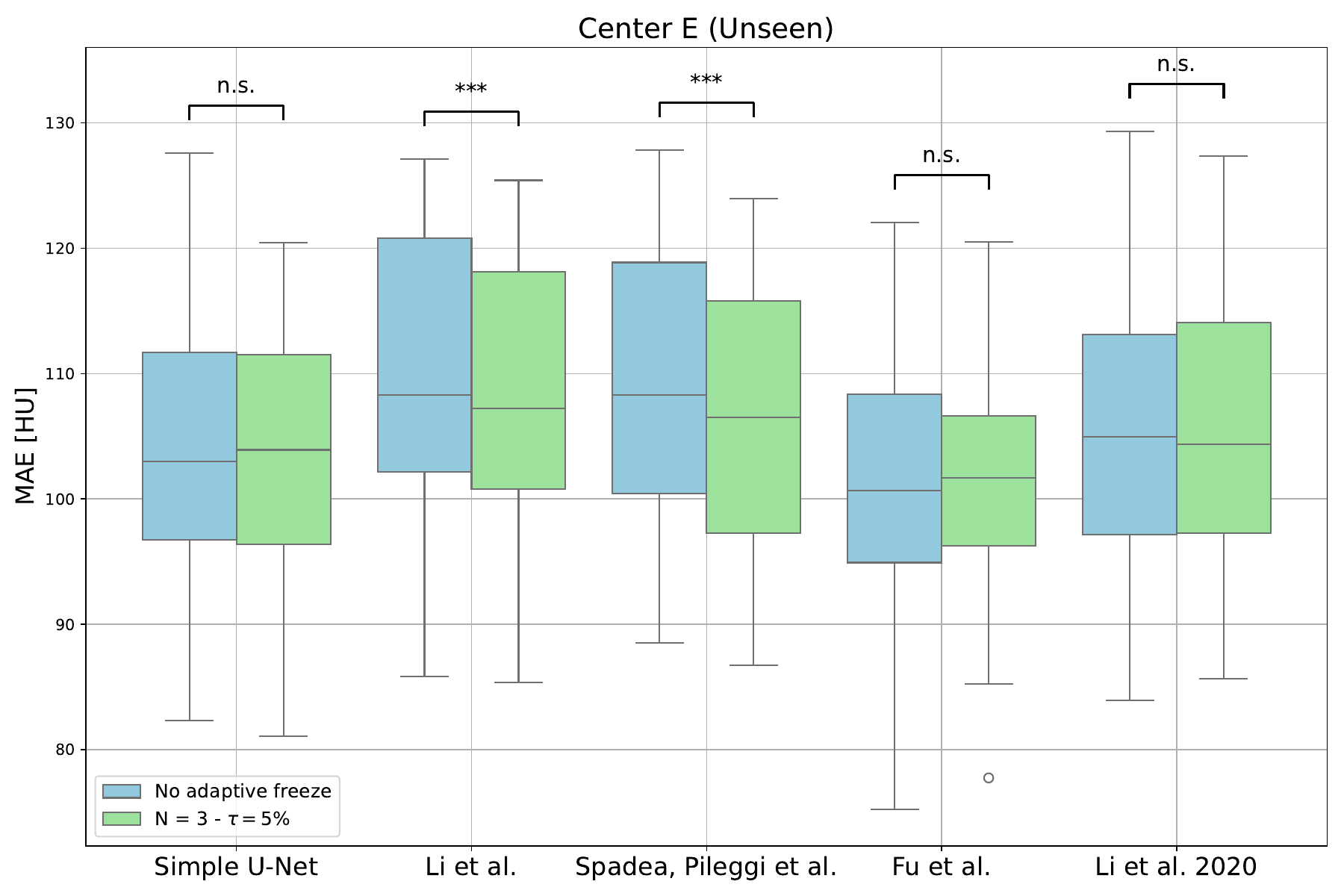}
\caption{Performance comparison of the models trained with and without the proposed patience-based adaptive encoder freezing. A $p-value < 0.001$ is indicated by $***$.} \label{fig:mae_comp}
\end{figure}

Nevertheless, the box-plot comparison of the MAE results obtained on the unseen Center E, for each model, is presented in Figure \ref{fig:mae_comp}. The statistical analysis indicated that for three of the five architectures (Simple U-Net \cite{UNetRonnenberger}, Fu et al. \cite{FuetalUNet2019}, and Li et al. \cite{LiUNet2020}) no statistically significant difference in MAE was found ($p>0.05$). In contrast, Li et al. \cite{Li2019} and Spadea, Pileggi et al. \cite{SPADEA2019495} demonstrated statistically significant improvements ($p<0.001$) in terms of MAE.

\begin{table}[ht]
    \centering
    \caption{Comparison of PSNR and SSIM results, in terms of median and interquartile range for each model and methodology.}
    \label{tab:psnr_ssim_results}
    \renewcommand{\multirowsetup}{\centering} 
    \begin{tabular}{llcc}
        \toprule
        \textbf{Model} & \textbf{Metric} & \textbf{\makecell{No Adaptive\\Freeze}} & \textbf{\makecell{Adaptive Freeze\\$\mathcal{N}=3, \space \tau=5\%$}} \\
        \midrule
        \multirow{2}[1]{*}{Simple U-Net \cite{UNetRonnenberger}} 
        & PSNR & 26.52 (25.49 - 27.36) & 26.48 (25.47 - 27.43)\\
        & SSIM & 0.89 (0.86 - 0.89) & 0.89 (0.86 - 0.89)\\
        \midrule
        \multirow{2}[1]{*}{Li et al. \cite{Li2019}} 
        & PSNR & 26.33 (25.32 - 27.18) & 26.42 (25.39 - 27.20)\\
        & SSIM & 0.89 (0.86 - 0.89) & 0.880 (0.86 - 0.89)\\
        \midrule
        \multirow{2}[1]{*}{Spadea, Pileggi et al. \cite{SPADEA2019495}} 
        & PSNR & 26.49 (25.38 - 27.37) & 26.43 (25.40 - 27.36)\\
        & SSIM & 0.89 (0.86 - 0.89) & 0.89 (0.86 - 0.89)\\
        \midrule
        \multirow{2}[1]{*}{Fu et al. \cite{FuetalUNet2019}} 
        & PSNR & 26.62 (25.53 - 27.45) & 26.59 (25.56 - 27.48)\\
        & SSIM & 0.89 (0.87 - 0.89) & 0.89 (0.86 - 0.89)\\
        \midrule
        \multirow{2}[1]{*}{Li et al, 2020 \cite{LiUNet2020}} 
        & PSNR & 26.41 (25.44 - 27.30) & 26.39 (25.40 - 27.30)\\
        & SSIM & 0.89 (0.86 - 0.89) & 0.89 (0.86 - 0.89)\\
        \bottomrule
    \end{tabular}
\end{table}

The PSNR and SSIM metrics are reported in Table \ref{tab:psnr_ssim_results}. No systematic differences between the two training strategies were observed, with only minor stochastic fluctuations across the evaluated architectures.

\section{Discussion}
This study introduces an adaptive encoder layer freezing strategy within federated encoder–decoder architectures for MRI-to-CT translation, operationalising the principle of equity through a Green AI by design approach.

The proposed strategy was deliberately confined to the encoder layers. As demonstrated and discussed in our previous study~\cite{raggio2025fedsynthct}, although federated models tend to generalise to unseen data, they exhibit reduced specificity at the client level. Extending the freezing strategy to the decoder would exacerbate this limitation, further restricting the capacity of the global model to accommodate local data heterogeneity. By focusing on the encoder, the federated model can instead exploit inter-client consensus to enhance stability and efficiency, while preserving decoder flexibility for client-specific adaptation. This design therefore achieves a balance between computational efficiency, convergence stability, and personalization.

To demonstrate its effectiveness, we evaluated the method on multiple encoder-decoder architectures.

In consideration of the results presented in Table \ref{tab:comparative_results}, on the Simple U-Net architecture by Ronneberger et al. \cite{UNetRonnenberger}, the adaptive freezing configuration (with parameters $\mathcal{N}=3$ and $\tau = 5\%$) resulted in a global reduction in training time of $16.8\%$ (from $13.40 \pm 0.76$ hours to $11.15 \pm 0.64$ hours). This temporal efficiency translated directly into proportional reductions in energy consumption and environmental impact, with total energy decreasing from $5.32 \pm 0.23$ kWh to $4.45 \pm 0.23$ kWh ($16.4\%$ reduction) and kgCO\textsubscript{2}eq emissions decreasing from $2.03 \pm 0.09$ kg to $1.69 \pm 0.09$ kg ($16.7\%$ reduction).

Improvements were observed with the Li et al. \cite{Li2019} architecture, where the adaptive freezing mechanism achieved a remarkable $22.0\%$ reduction in training time (from $10.11 \pm 0.38$ hours to $7.89 \pm 0.22$ hours). Correspondingly, this architecture exhibited the most substantial environmental benefits, with kgCO\textsubscript{2}eq emissions reduced by $23.2\%$ (from $1.51 \pm 0.04$ kg to $1.16 \pm 0.03$ kg) and energy consumption decreased by $23.0\%$ (from $3.96 \pm 0.10$ kWh to $3.05 \pm 0.07$ kWh). 

The positive trend was confirmed for the Fu et al. \cite{FuetalUNet2019} architecture, which exhibited substantial efficiency improvements, with training time reduced by 16.2\% (from $16.49 \pm 0.57$ hours to $13.81 \pm 0.64$ hours). This architecture also demonstrated notable environmental benefits, with kgCO\textsubscript{2}eq emissions decreasing by 17.1\% (from $2.17 \pm 0.07$ kg to $1.80 \pm 0.07$ kg) and energy consumption reduced by 16.7\% (from $5.69 \pm 0.17$ kWh to $4.74 \pm 0.20$ kWh).

Similarly, the Li et al. 2020 \cite{LiUNet2020} architecture showed consistent improvements with a 16.6\% reduction in training time (from $14.10 \pm 0.28$ hours to $11.76 \pm 0.59$ hours), as well as corresponding decreases in emissions (16.8\%, from $1.85 \pm 0.03$ kg to $1.54 \pm 0.07$ kg) and energy consumption (16.7\%, from $4.85 \pm 0.09$ kWh to $4.04 \pm 0.19$ kWh).

Notably, the Spadea, Pileggi et al.\cite{SPADEA2019495} architecture, despite requiring the longest absolute training time, demonstrated consistent improvements with a 9.4\% reduction in training duration (from $19.22 \pm 0.75$ hours to $17.42 \pm 0.68$ hours) across 25 rounds, with a corresponding reductions in emissions and energy consumption of $\approx9\%$. 

Furthermore, Figure~\ref{fig:time_client_comp_pre_post} and Table~\ref{tab:training_times} report and compare the training time per local epoch before and after the application of the proposed methods across the four clients.

As a direct consequence of the reduction in the number of the trainable parameters due to the encoder freeze, in Centre A, the Simple U-Net average time decreased from 21.07 $\pm$ 0.65 min to 14.38 $\pm$ 0.43 min per local epoch, corresponding to a 31.8\% reduction, while Li et al.\cite{Li2019} reduced from 15.59 $\pm$ 0.53 min to 10.05 $\pm$ 0.31 min (35.5\% reduction). Similar reduction trends were observed in Centres B, C, and D, with the largest absolute reductions observed in Spadea, Pileggi et al. on Centre D (from 44.03 $\pm$ 0.79 min to 35.77 $\pm$ 0.59 min) and Fu et al. on Centre D (from 38.50 $\pm$ 1.90 min to 27.71 $\pm$ 1.35 min).

These per-epoch reductions scaled across the full federated training process, leading to proportional decreases in total energy consumption and CO\textsubscript{2} emissions observed in Table \ref{tab:comparative_results}, highlighting that adaptive freezing not only accelerates training but also improves the environmental efficiency of the federated learning workflow, without adversely affecting model performance across heterogeneous client datasets.

These findings demonstrated the efficacy of the patience-based adaptive freezing approach in a federated environment, with consistent improvements observed across various architectures, ranging from fundamental designs to more complex variants.

The statistical analysis of the models' performance -- in terms of MAE -- on the unseen Centre E dataset (Figure \ref{fig:mae_comp}) revealed that the proposed adaptive freezing mechanism preserved generalisation performance and occasionally marginally enhanced it. 

For three of the five evaluated architectures (Simple U-Net \cite{UNetRonnenberger}, Fu et al. \cite{FuetalUNet2019}, and Li et al. \cite{LiUNet2020}), no statistically significant difference in MAE was found ($p > 0.05$, t-test) on 23 unseen patients, indicating that computational efficiency gains were achieved without compromising model performance. The conduction of statistical testing across single clients was precluded by the limited number of test cases available per site.

Notably, the results for the Li et al. \cite{Li2019} and Spadea, Pileggi et al. \cite{SPADEA2019495} architectures, demonstrated statistically significant improvements in terms of MAE ($p < 0.001$, t-test). However, the absolute MAE reduction ($\approx$2-3 HU) remains well below radiologically and clinically meaningful thresholds for image contrast and radiotherapy, where deviations greater than 20-50 HU are typically required to discriminate biological tissues and affect dose distributions~\cite{Liu2016-ng, Davis2019-ru}. Hence, these results indicated that the proposed adaptive freezing mechanism preserved model fidelity, with potential minor regularisation effects, but without producing relevant performance changes.

Indeed, no performance improvement or deterioration was observed in terms of PSNR and SSIM -- metrics that are less sensitive than MAE -- with only minor stochastic fluctuations resulting from the independent repetitions of the experiments. Comparable or identical findings across the different architectures and methodologies was also reported in the FedSynthCT-Brain study \cite{raggio2025fedsynthct}, particularly with regard to SSIM, further emphasising that MAE remains the pivotal metric for performance evaluation in this type of task.

To further investigate the impact of the adaptive freezing mechanism on the generated sCTs, additional qualitative comparisons are provided in the supplementary material (see Section S.1, Figure S1-S10). Specifically, for each evaluated model, two representative cases are presented: (i) the case with the smallest difference and (ii) the case with the largest difference in MAE between the sCTs generated by the models trained with and without adaptive freezing, across the test set.

The visual comparisons confirmed that no systematic degradations are associated with the proposed methodology, and the observed differences were attributable to the inherent stochastic nature of the training process and the independent experiments.

The standard deviations reported for each metric across five repetitions indicate stable and reproducible results, with the adaptive freezing methodology consistently reducing variability in training time and resource consumption. The emissions reductions ranged from 9.1\% to 23.2\%, with an average reduction of $\approx19.9\%$ across all models. These improvements are particularly important given the increasing emphasis on sustainable machine learning practices and the growing concern regarding the carbon footprint of DL model training \cite{DELANOE2023117261}. Furthermore, this efficiency gain is especially valuable in clinical settings where computational resources may be limited and cost considerations are essential.

The parameters $\mathcal{N}=3$ and $\tau = 5\%$ were found to provide an optimal balance between computational efficiency and performance preservation in the evaluated architectures, and for the MRI-to-CT translation task. While these values proved to be flexible and effective across the five encoder-decoder architectures considered, they may not be optimal for other encoder-decoder architectures. Establishing suitable values requires initially observing the evolution of the relative percentage difference in the aggregated encoder ($\rho_{\%}$) without freezing across training rounds, as represented in Figure \ref{fig:method}. This procedure offers a principled methodology for selecting hyperparameters that balance client-specific adaptability, convergence stability and computational efficiency.

Consequently, the presented configuration should be regarded as a reasonable baseline, while systematic tuning would be necessary for other encoder-decoder-based architectures, such as GAN generators.

\section{Conclusion}
In this study we laid the foundation for operationalising the principle of equity in FL by applying the Green AI's theory by design. We achieved this by introducing a patience-based adaptive freezing strategy for encoder layers in federated encoder-decoder-based architectures for MRI-to-CT translation. By leveraging relative percentage differences in encoder weights, our approach effectively reduced computational demands, emissions, and total energy consumption across rounds. Unlike conventional layer freezing techniques, which primarily focus on communication and memory efficiency, our method integrates environmental, economic, and social considerations as well, with the aim of making the federated strategy more accessible to a wider range of users in the medical field. The required manual tuning required for the $\tau$ and patience parameters might represent a limitation of this methodology, because it may necessitate some preliminary investigation of the specific model in use, in order to observe the dynamics of weight variations and to select appropriate values. As a direction for future work, it would be valuable to explore automated strategies for determining the correct $\tau$ threshold or to develop a systematic protocol to guide the selection of both $\tau$ and patience parameters, thereby enhancing the practicality and generalisability of the approach.

\section*{Data availability statement}
The data used for the study acquired from Centre A, B and C are private. The data from Centre D and Centre E were extracted from the public SynthRAD2023 Grand Challenge dataset and are available at \url{https://doi.org/10.5281/zenodo.7260705}.

\section*{Funding}
This research did not receive any specific grant from funding agencies in the public, commercial, or not-for-profit sectors.

\section*{Ethics statement}
Written informed consent was obtained from all participants. The study complied with the ethical standards of the 1964 Declaration of Helsinki and its later amendments for research involving human subjects.
All patient data were fully anonymized prior to analysis to ensure privacy and confidentiality. No personally identifiable information was used in this study.

\section*{Declaration of competing interest}
The authors declare that they have no known competing financial interests or personal relationships that could have appeared to influence the work reported in this paper.

\newpage

\section*{Supplementary Material}

\setcounter{figure}{0}

\renewcommand{\thefigure}{S\arabic{figure}}

\subsection*{S.1 - Visual Comparison of Generated sCTs}

To qualitatively assess the potential impact of the adaptive freezing strategy on the generated synthetic CTs (sCTs), we conducted a visual comparison between the sCTs obtained with and without the proposed methodology. For each model employed in the study, we selected two representative test cases based on the following criteria:
\begin{itemize}
    \item The case with the \textbf{minimum difference in terms of MAE}, thus the lowest difference between the sCTs MAE generated with and without adaptive freezing.
    \item The case with the \textbf{maximum difference in terms of MAE}, thus the highest difference between the sCTs MAE generated with and without adaptive freezing.
\end{itemize}

\noindent Each figure presents the following for the axial, coronal and sagittal plane of the central slice:
\begin{enumerate}
    \item Input MR image;
    \item Ground-truth CT image;
    \item sCT generated without adaptive freezing;
    \item sCT generated with the proposed adaptive freezing, using $\mathcal{N}=3$ and $\tau=5\%$;
    \item Absolute difference map between the two sCTs.
\end{enumerate}

\begin{figure}[H]
    \centering
    \includegraphics[width=\textwidth]{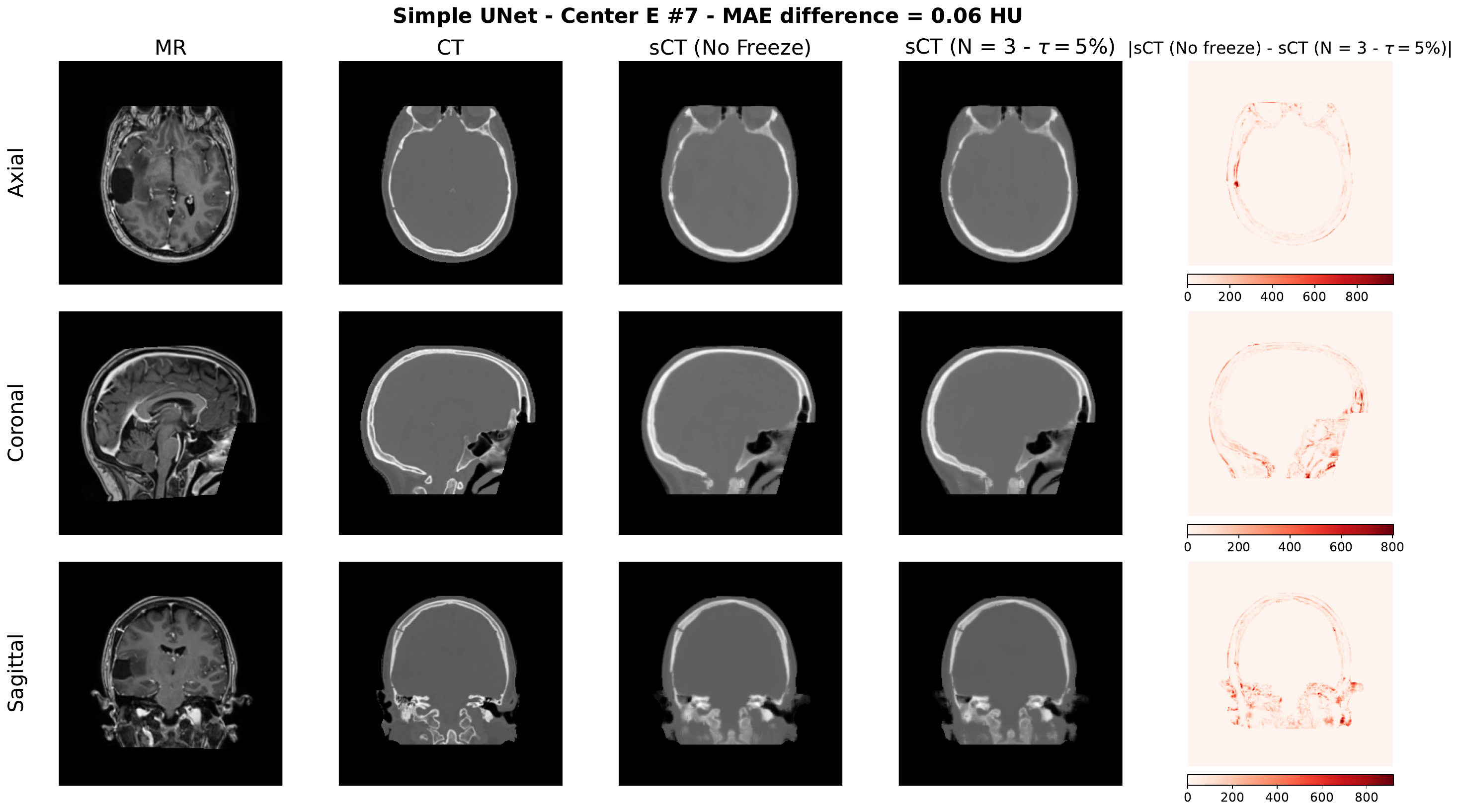}
    \caption{Simple UNet \cite{UNetRonnenberger} – Minimum MAE difference case.}
    
\end{figure}

\begin{figure}[H]
    \centering
    \includegraphics[width=\textwidth]{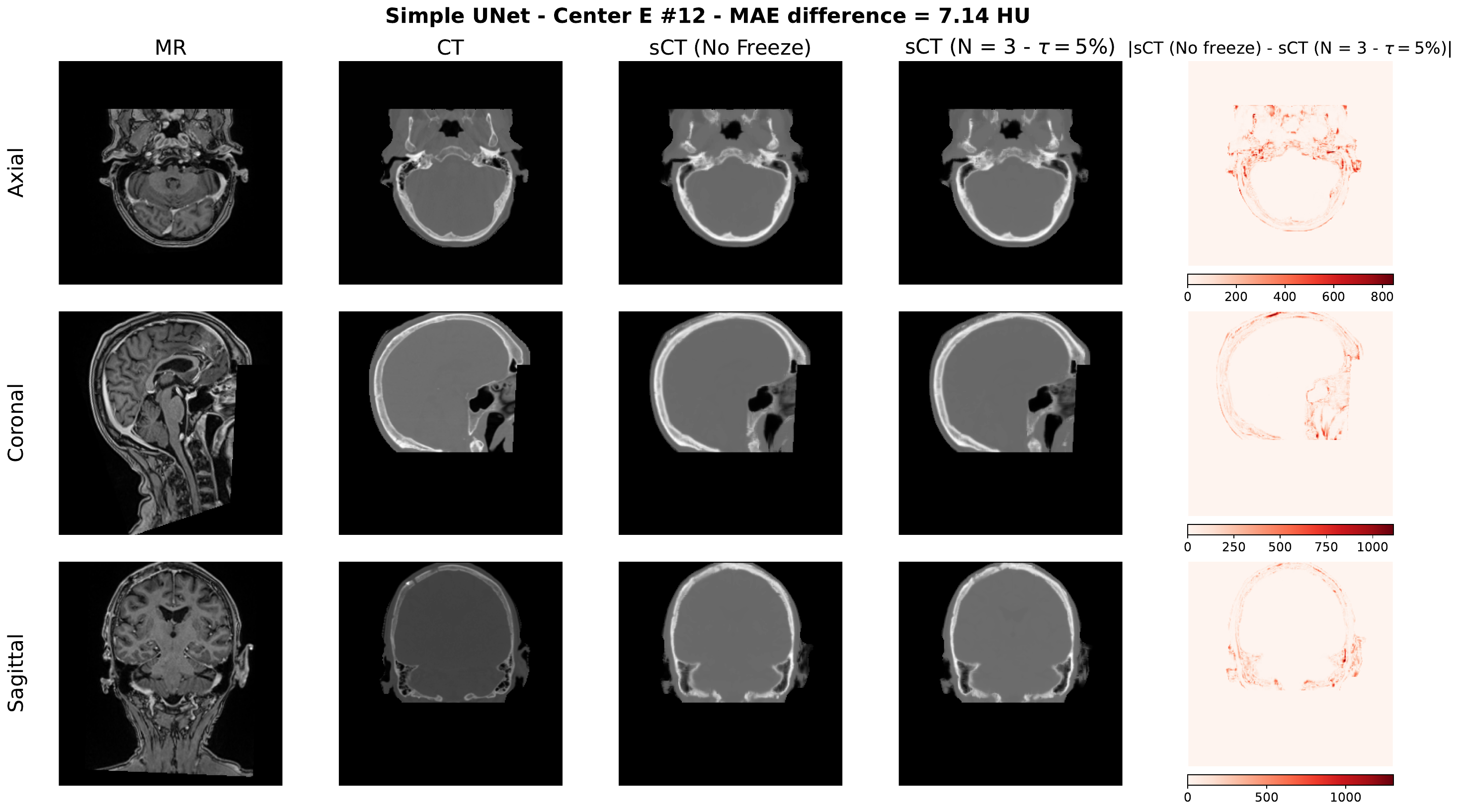}
    \caption{Simple UNet \cite{UNetRonnenberger} – Maximum MAE difference case.}
    
\end{figure}
\clearpage

\begin{figure}[H]
    \centering
    \includegraphics[width=\textwidth]{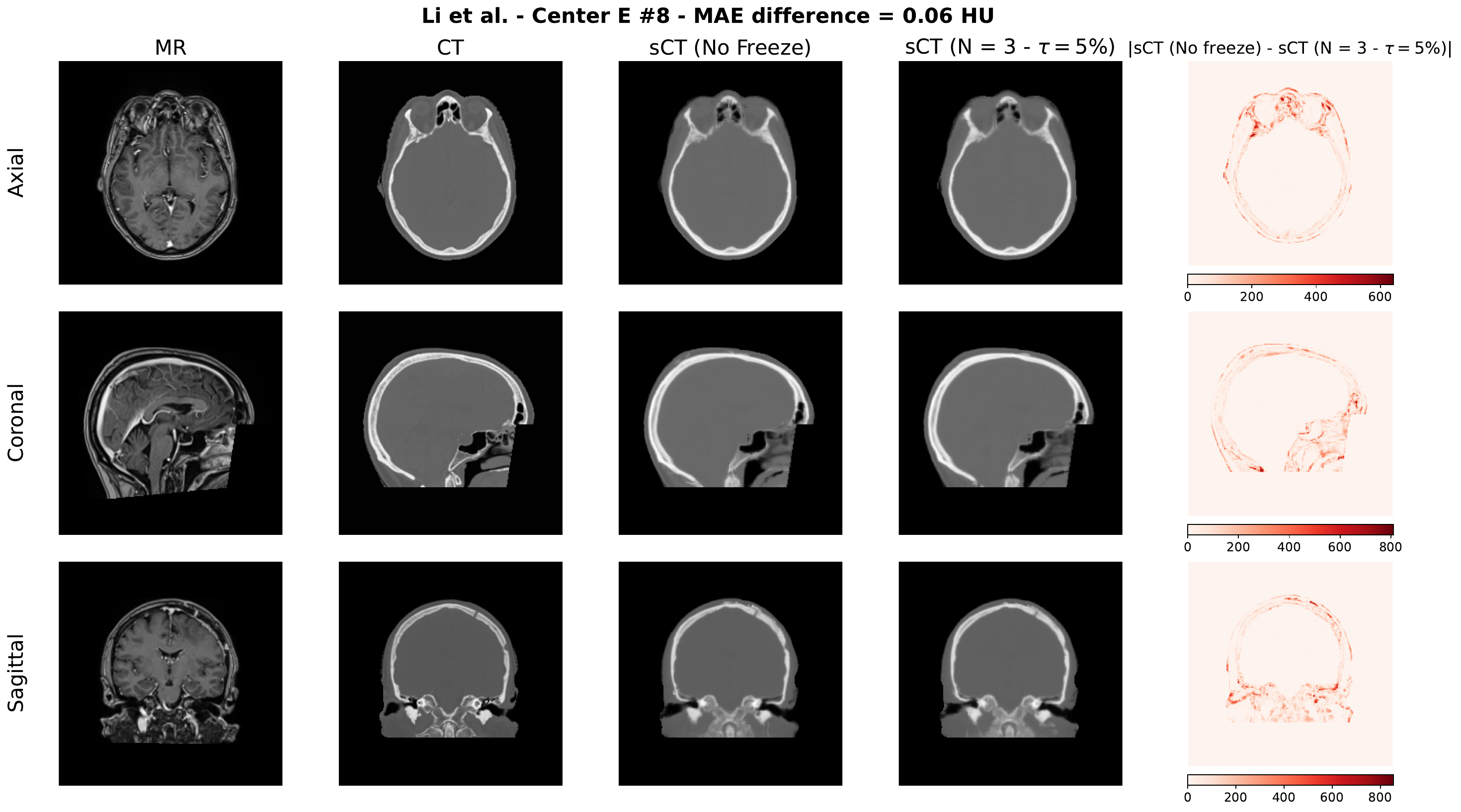}
    \caption{Li et al. architecture  \cite{Li2019} – Minimum MAE difference case.}
    
\end{figure}

\begin{figure}[H]
    \centering
    \includegraphics[width=\textwidth]{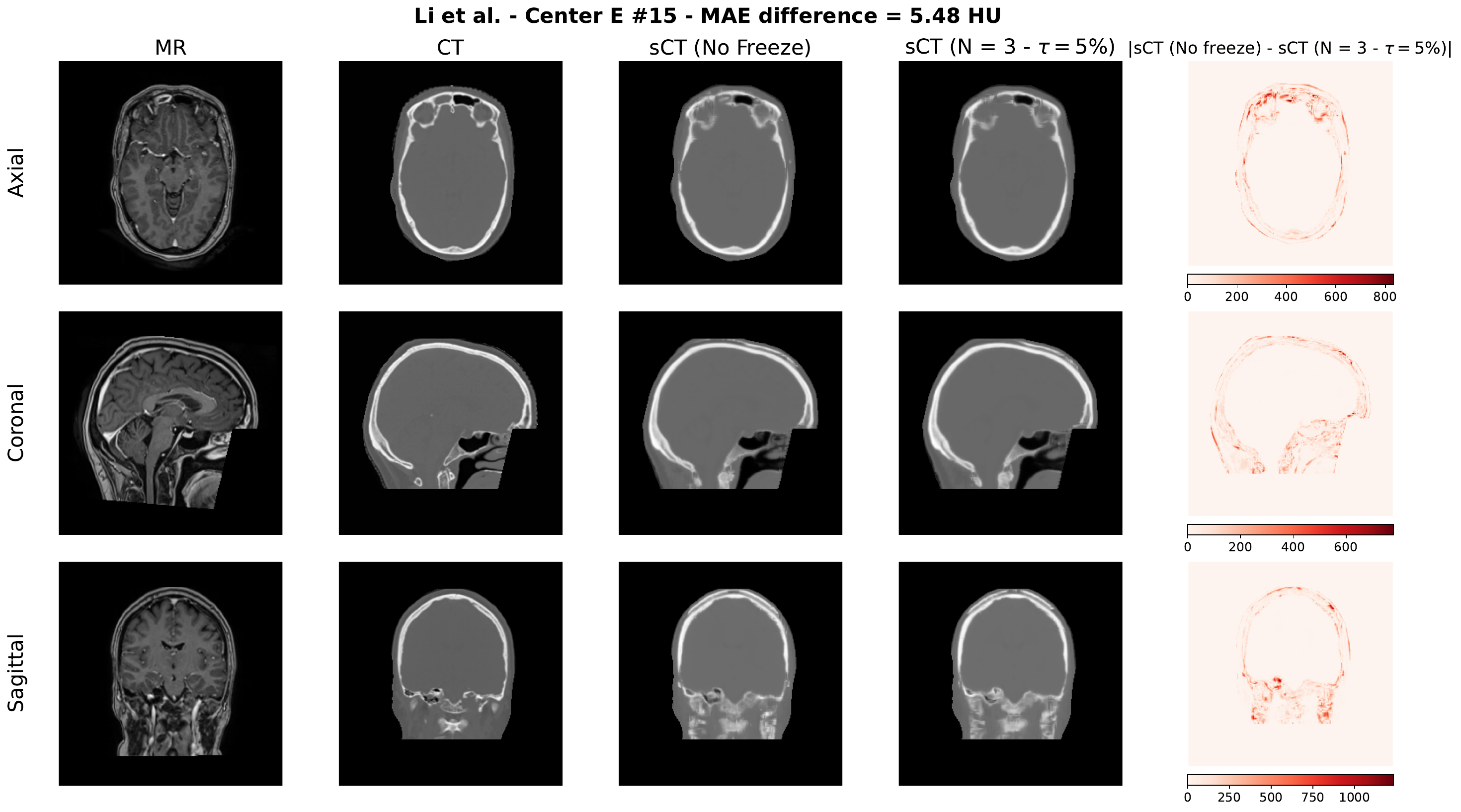}
    \caption{Li et al. architecture \cite{Li2019} – Maximum MAE difference case.}
    
\end{figure}
\clearpage

\begin{figure}[H]
    \centering
    \includegraphics[width=\textwidth]{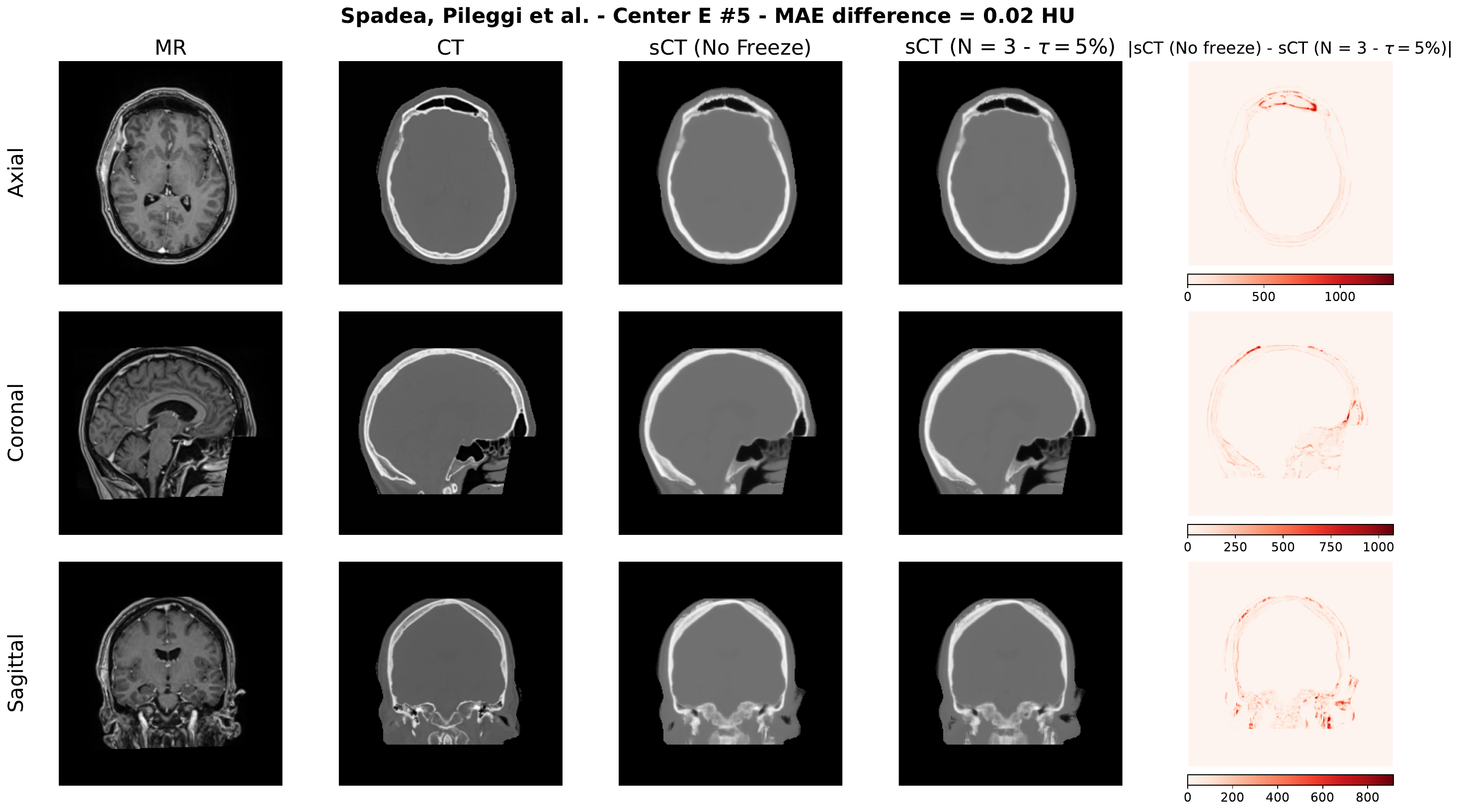}
    \caption{Spadea, Pileggi et al. architecture \cite{SPADEA2019495} – Minimum MAE difference case.}
    
\end{figure}

\begin{figure}[H]
    \centering
    \includegraphics[width=\textwidth]{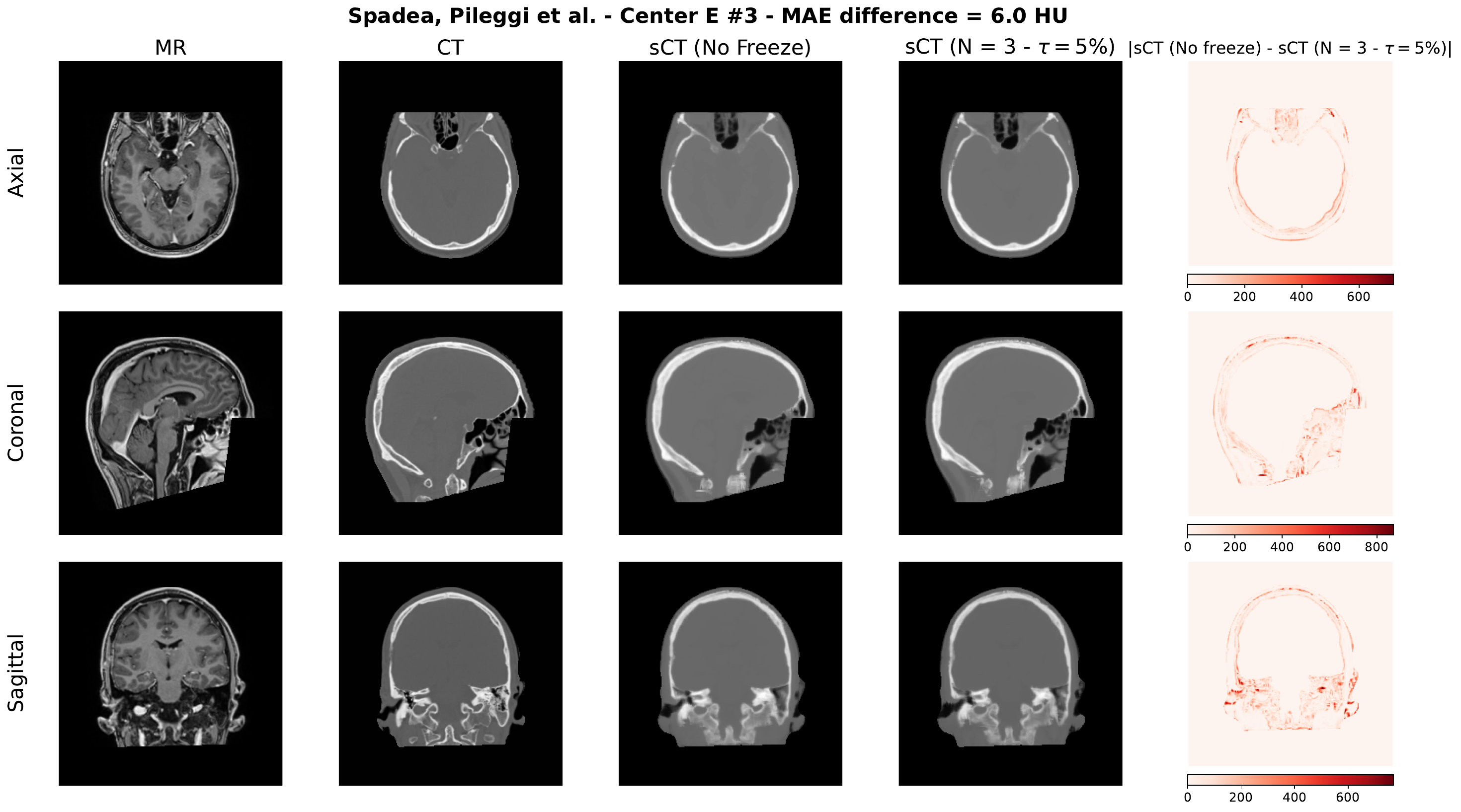}
    \caption{Spadea, Pileggi et al. architecture \cite{SPADEA2019495} – Maximum MAE difference case.}
    
\end{figure}

\begin{figure}[H]
    \centering
    \includegraphics[width=\textwidth]{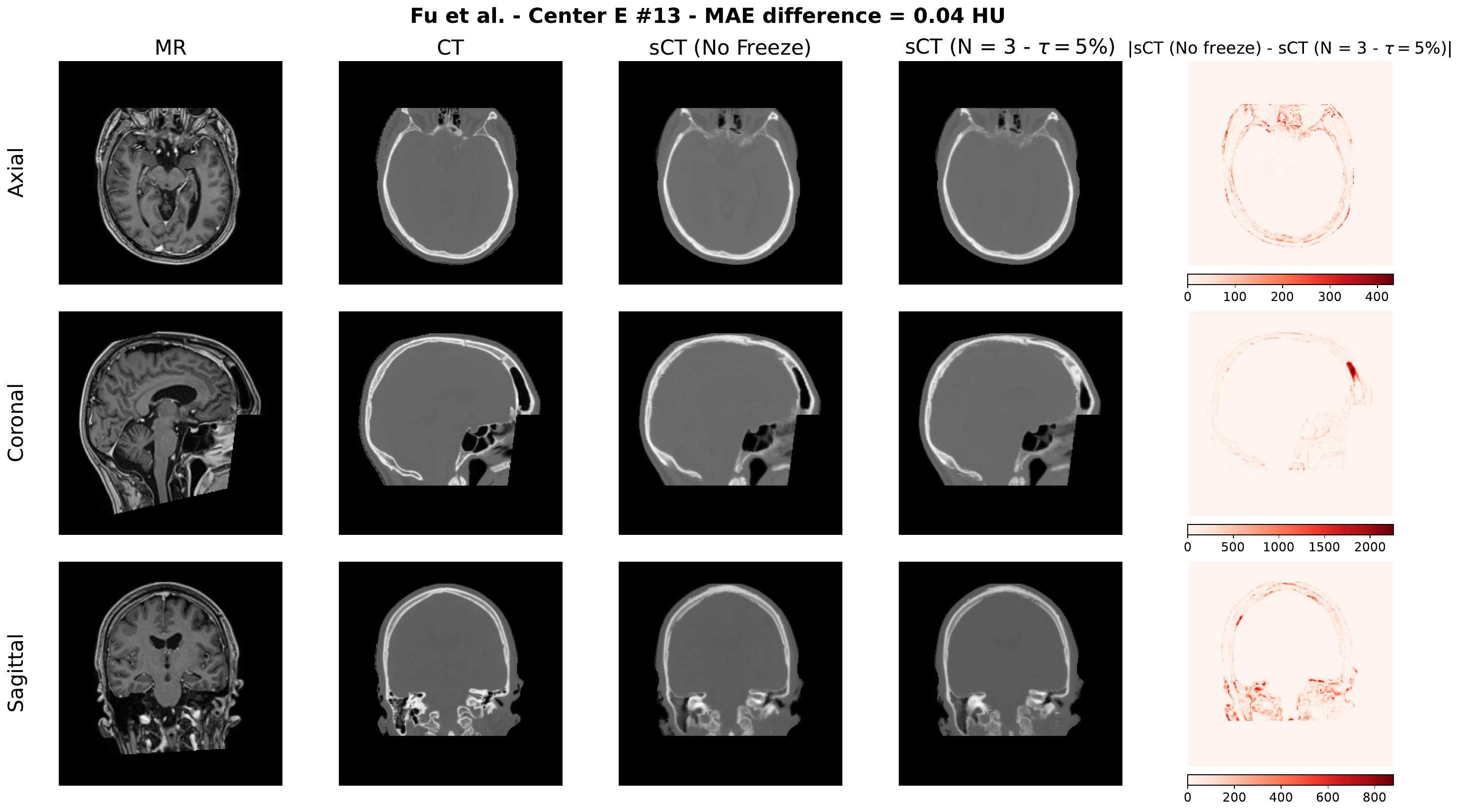}
    \caption{Fu et al. architecture \cite{FuetalUNet2019} – Minimum MAE difference case.}
    
\end{figure}

\begin{figure}[H]
    \centering
    \includegraphics[width=\textwidth]{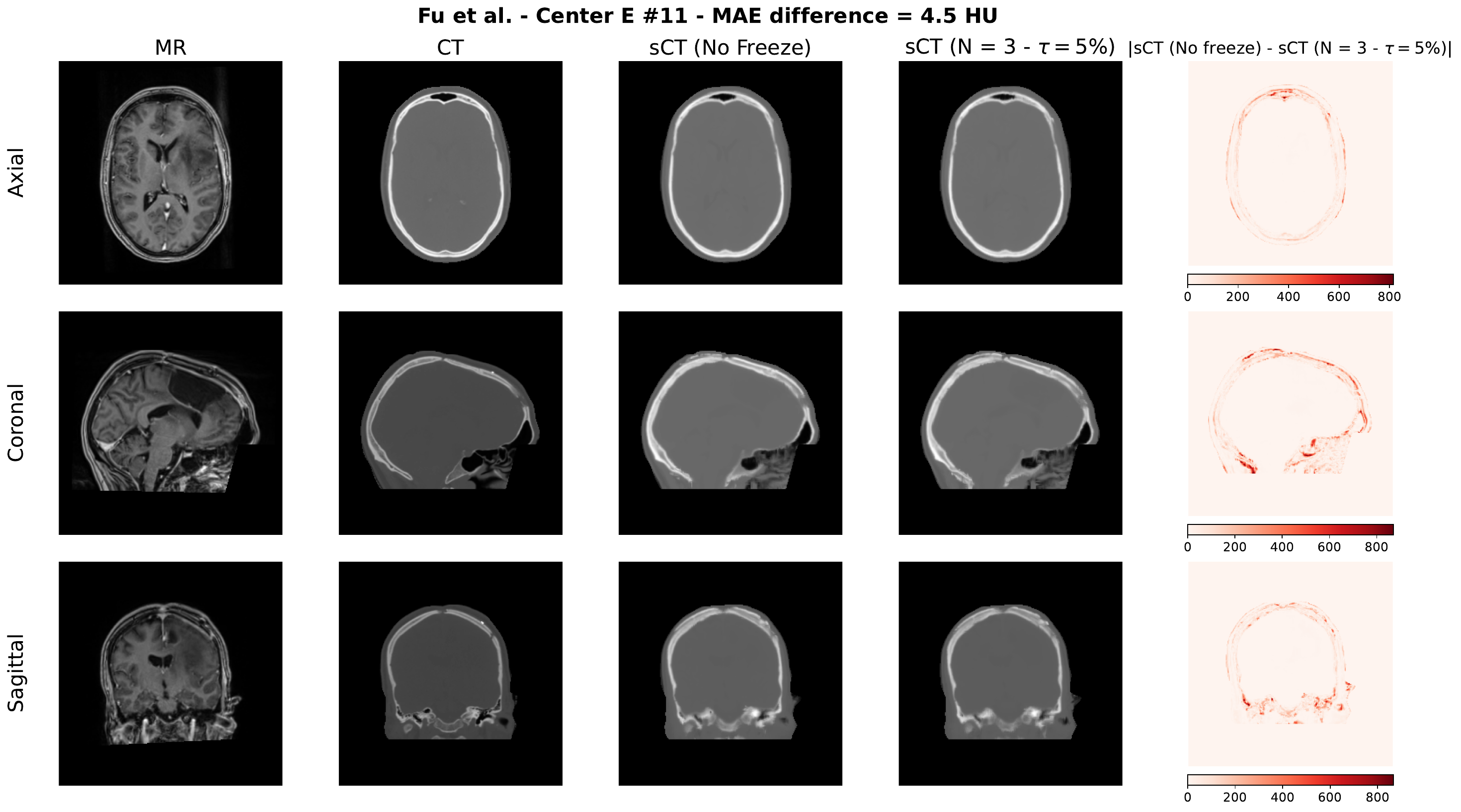}
    \caption{Fu et al. architecture \cite{FuetalUNet2019} – Maximum MAE difference case.}
    
\end{figure}

\begin{figure}[H]
    \centering
    \includegraphics[width=\textwidth]{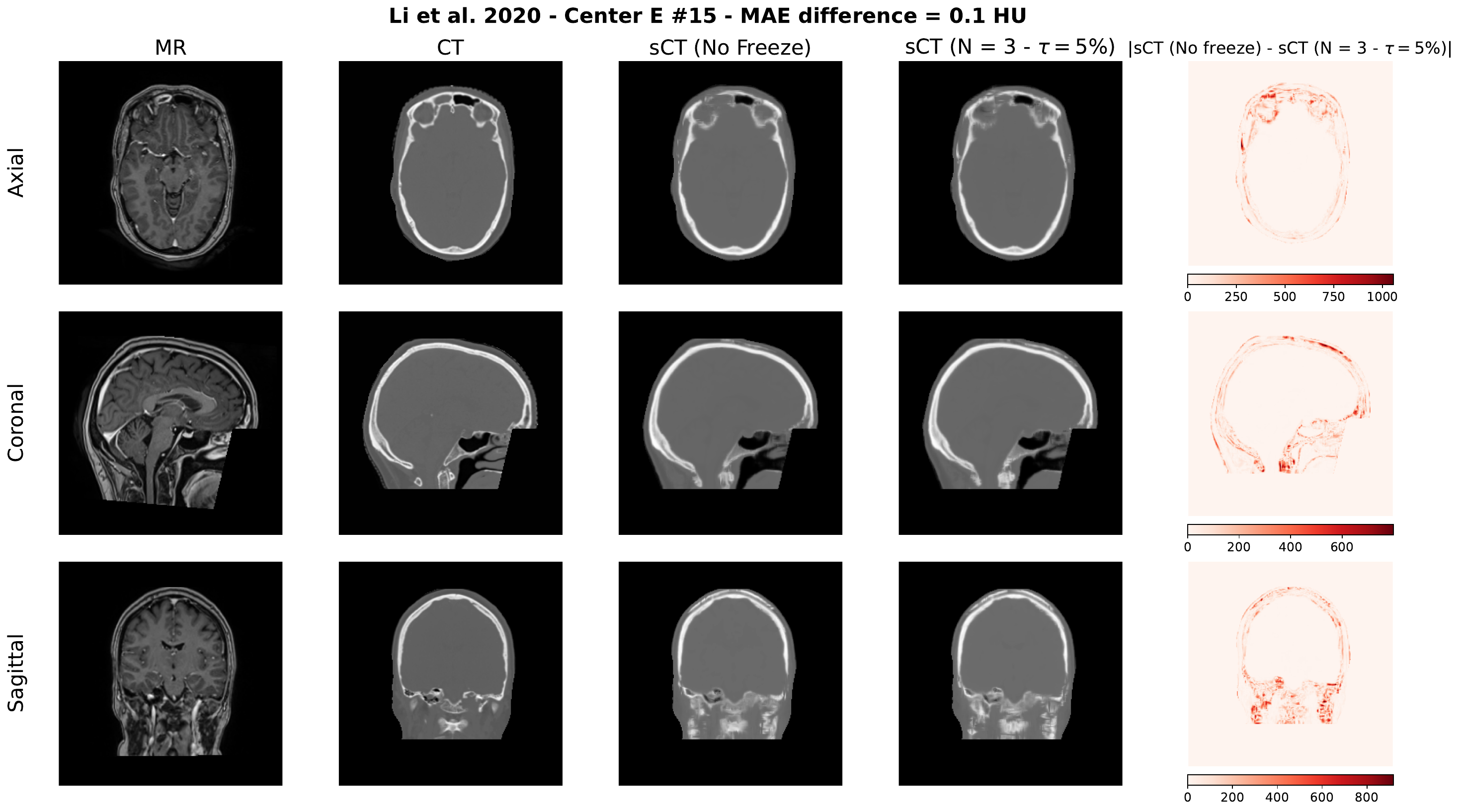}
    \caption{Li et al. 2020 architecture \cite{LiUNet2020} – Minimum MAE difference case.}
    
\end{figure}

\begin{figure}[H]
    \centering
    \includegraphics[width=\textwidth]{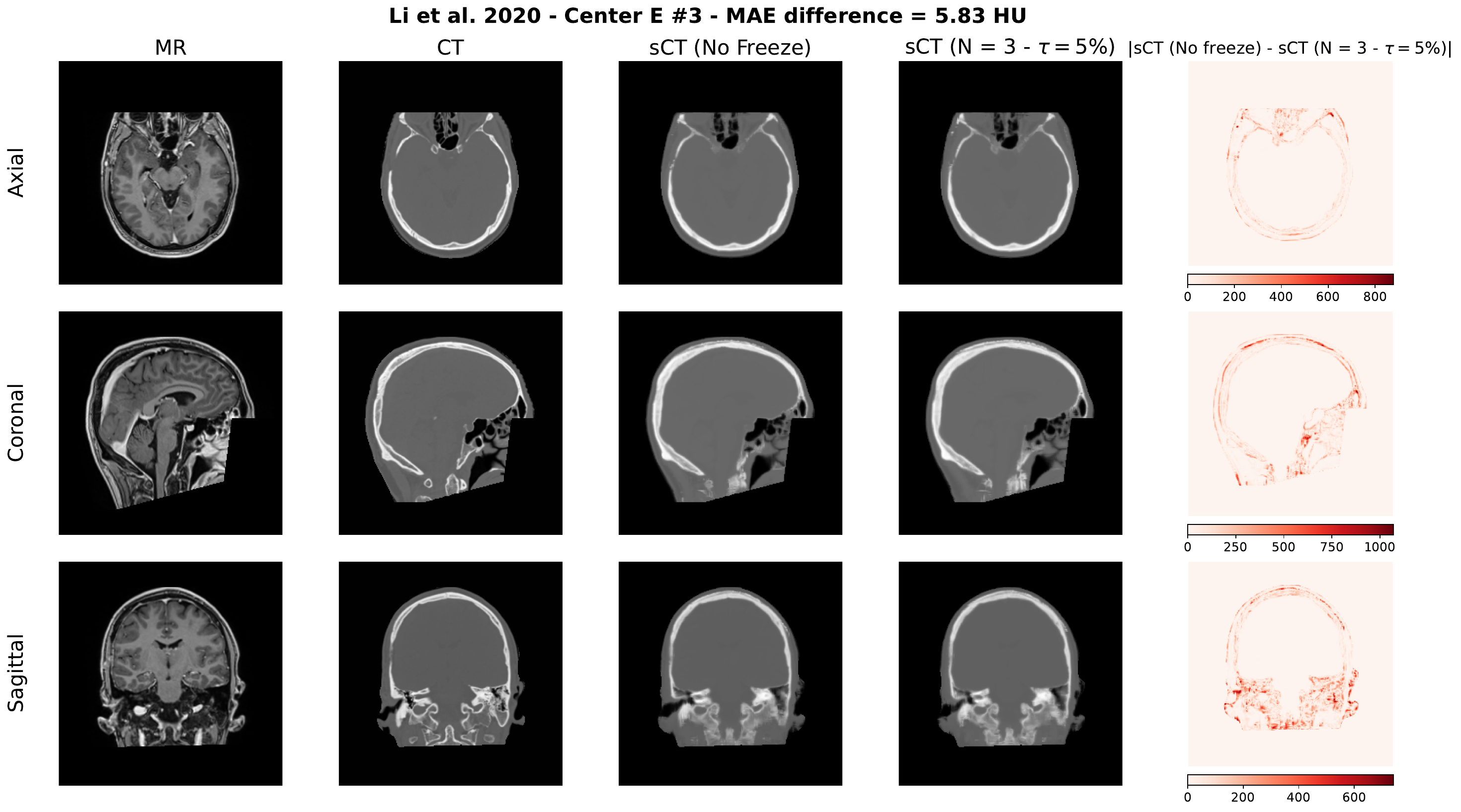}
    \caption{Li et al. 2020 architecture \cite{LiUNet2020} – Minimum MAE difference case.}
    
\end{figure}

Across all models and both selected cases, no relevant and systematic structural differences were observed. The differences highlighted in the absolute difference maps and the variation in MAE are attributable to the stochastic nature of the training process (e.g., batch shuffling and optimization path variability) rather than the direct effect of the adaptive freezing mechanism.

These findings are aligned with the results observed in terms of MAE, PSNR and SSIM presented in this study.

\bibliographystyle{unsrt}

\end{document}